\documentclass[11pt]{article}

\usepackage{PRIMEarxiv}

\usepackage[utf8]{inputenc} 
\usepackage[T1]{fontenc}    
\usepackage{microtype}      

\usepackage{amsmath, amssymb} 
\usepackage{amsfonts}       
\usepackage{nicefrac}       

\usepackage{booktabs}       
\usepackage{graphicx}       
\usepackage{subcaption}     
\usepackage{float}          
\usepackage{adjustbox}      
\usepackage{caption}        
\usepackage{algorithm}
\usepackage{algpseudocode}

\usepackage[authoryear]{natbib} 
\usepackage{hyperref}       
\hypersetup{colorlinks=true, linkcolor=blue, citecolor=blue, urlcolor=blue}
\usepackage{url}            

\usepackage{fancyhdr}       
\usepackage{enumitem}
\setlist[itemize]{leftmargin=*}
\usepackage{xcolor}
\usepackage{tikz} 
\usetikzlibrary{shapes.geometric, arrows.meta, positioning}
\usepackage{rotating} 
\graphicspath{{media/}} 

\title{NIRMAL Pooling: An Adaptive Max Pooling Approach with Non-linear Activation for Enhanced Image Classification}

\usepackage{authblk} 

\author[1]{Nirmal Gaud}
\author[2]{Krishna Kumar Jha}
\author[3]{Jhimli Adhikari}
\author[4]{Adhini Nasarin P S}
\author[5]{Joydeep Das}
\author[6]{Samarth S Deshpande}
\author[7]{Nitasha Barara}
\author[8]{Vaduguru Venkata Ramya}
\author[9]{Santu Saha}
\author[10]{Mehmet Tarik Baran}
\author[11]{Sarangi Venkateshwarlu}
\author[12]{Anusha M D}
\author[13]{Surej Mouli}
\author[14]{Preeti Katiyar}
\author[15]{Vipin Kumar Chaudhary}

\affil[1]{CEO and Master Trainer, ThinkAI - A Machine Learning Community}
\affil[2]{Guru Nanak Institute of Technology, India}
\affil[3]{Department of Computer Science, Narayan Zantye College, Goa, India}
\affil[4]{NIIT Foundation, India}
\affil[5]{Not affiliated}
\affil[6]{Dayananda Sagar College of Engineering, Bangalore, India}
\affil[7]{IILM, Lodhi Road, India}
\affil[8]{Sri Balaji University, Pune, India}
\affil[9]{University of Calcutta, India}
\affil[10]{Mardin Artuklu University, Turkey}
\affil[11]{City University of Hong Kong, Hong Kong}
\affil[12]{Yenepoya (Deemed to be University), India}
\affil[13]{Aston University, Birmingham, UK}
\affil[14]{Delhi Technical Campus, Greater Noida, India}
\affil[15]{Lovely Professional University, Phagwara, Punjab, India}

\begin{document}
\maketitle

\begin{abstract}
This paper presents NIRMAL Pooling, a novel pooling layer for Convolutional Neural Networks (CNNs) that integrates adaptive max pooling with non-linear activation function for image classification tasks. The acronym NIRMAL stands for Non-linear Activation, Intermediate Aggregation, Reduction, Maximum, Adaptive, and Localized. By dynamically adjusting pooling parameters based on desired output dimensions and applying a Rectified Linear Unit (ReLU) activation post-pooling, NIRMAL Pooling improves robustness and feature expressiveness. We evaluated its performance against standard Max Pooling on three benchmark datasets: MNIST Digits, MNIST Fashion, and CIFAR-10. NIRMAL Pooling achieves test accuracies of 99.25\% (vs. 99.12\% for Max Pooling) on MNIST Digits, 91.59\% (vs. 91.44\%) on MNIST Fashion, and 70.49\% (vs. 68.87\%) on CIFAR-10, demonstrating consistent improvements, particularly on complex datasets. This work highlights the potential of NIRMAL Pooling to enhance CNN performance in diverse image recognition tasks, offering a flexible and reliable alternative to traditional pooling methods.
\end{abstract}


\section{Introduction}
Convolutional Neural Networks have revolutionized the field of computer vision by achieving remarkable success in image processing tasks such as classification, object detection, and segmentation \citep{krizhevsky2012imagenet}. A fundamental building block of CNNs is the pooling layer, which performs dimensionality reduction, thus decreasing computational complexity and reducing overfitting. Traditional pooling methods, such as Max Pooling and Average Pooling, rely on fixed window sizes and strides, which may not optimally adapt to varying input characteristics or task complexities.

To address these limitations, we propose NIRMAL Pooling, a novel pooling layer that integrates adaptive pooling with a non-linear activation function. The acronym NIRMAL encapsulates its core features: Non-linear Activation, Intermediate Aggregation, Reduction, Maximum, Adaptive, and Localized. Unlike traditional pooling, NIRMAL Pooling dynamically calculates window sizes and strides based on desired output dimensions, followed by a ReLU activation to enhance feature expressiveness. This approach aims to improve the representation of features and robustness of the model, particularly for complex datasets like CIFAR-10 and CIFAR-100.

This paper provides a comprehensive exploration of NIRMAL Pooling, including its algorithmic design, mathematical formulation, implementation details, and experimental evaluation. We compare its performance against standard Max Pooling across three benchmark datasets, demonstrating its efficiency and potential for broader applications. The paper is organized as follows: Section \ref{sec:literature} reviews related work on pooling methods, Section \ref{sec:acronym} explains the acronym NIRMAL, Section \ref{sec:algorithm} details the algorithm, Section \ref{sec:math} presents the mathematical formulation, Section \ref{sec:implementation} discusses implementation considerations, Section \ref{sec:results} presents experimental results, Section \ref{sec:discussion} analyzes the findings, and Section \ref{sec:future} outlines future research directions.

\section{Literature Review}
\label{sec:literature}
Pooling layers are integral to Convolutional Neural Networks. Recent research has explored innovative pooling methods to address the limitations of traditional approaches such as max pooling and average pooling. This section reviews five studies that propose or analyze novel pooling techniques, situating NIRMAL Pooling within this context.

\citep{rodriguez2022pooling} investigated alternative pooling operators for CNNs in the prediction of COVID-19 using chest X-ray images. They replaced traditional pooling layers with aggregation theory functions, demonstrating that these modifications lead to distinct model behaviors, particularly when prioritizing metrics such as precision or recall. Unlike NIRMAL Pooling, which integrates adaptive max pooling with non-linear activation, their approach focuses on application-specific performance tuning, highlighting the potential of customized pooling strategies.

\citep{akgul2025mam} proposed the Maximum Average Minimum (MAM) pooling method, which computes a representative pixel value by combining maximum, average, and minimum operations. Tested on LeNet-5 with MNIST, CIFAR-10, and CIFAR-100 datasets, MAM outperformed traditional pooling methods in various pool sizes. Although MAM emphasizes interactivity and reduced data loss, NIRMAL Pooling’s adaptive parameter calculation and ReLU activation offer a different approach to enhancing feature expressiveness, particularly for complex datasets like CIFAR-10.

\citep{zafar2024comparison} provided a comprehensive comparison of pooling methods, including Pyramid Pooling, Stochastic Pooling, and Fractional Max Pooling. They noted that Pyramid Pooling accommodates varying input sizes but is complex for deep networks, while Stochastic Pooling risks losing critical features due to its randomness. NIRMAL Pooling addresses these issues by dynamically adjusting parameters and incorporating non-linear activation, offering a balance of flexibility and robustness without the complexity of Pyramid Pooling or the unpredictability of Stochastic Pooling.

\citep{shadoul2022effect} explored the integration of learnable parameters into pooling layers to overcome the limitations of Max Pooling, which may discard useful information, and Average Pooling, which treats all pixels equally. Their findings suggest that learnable pooling kernels enhance adaptability and performance. NIRMAL Pooling shares the goal of adaptability but achieves it through dynamic window and stride calculations rather than trainable parameters, potentially reducing computational overhead while maintaining effectiveness.

\citep{zhao2022improved} introduced the T-Max-Avg pooling, which uses a threshold parameter to switch between maximum values and a weighted average of the top-K pixels. Evaluated on MNIST, CIFAR-10, and CIFAR-100, this method outperformed traditional pooling in LeNet-5 models by learning optimal pooling strategies during training. NIRMAL Pooling similarly aims to optimize feature selection but does so through adaptive max pooling and ReLU activation, offering a simpler yet effective alternative that avoids the need for threshold tuning.

These studies collectively underscore the importance of tailoring pooling methods to specific tasks and datasets. NIRMAL Pooling distinguishes itself by combining adaptive parameterization with non-linear activation, providing a flexible and robust solution that enhances feature representation without the complexity of learnable parameters or threshold-based mechanisms. Its performance on benchmark datasets, particularly CIFAR-10, aligns with the trend to develop task-adaptive pooling methods while offering a novel contribution through its ReLU-enhanced design.

\section{NIRMAL Acronym}
\label{sec:acronym}
The NIRMAL acronym encapsulates the key characteristics of the proposed pooling layer:
\begin{itemize}
    \item \textbf{Non-linear Activation}: A ReLU function is applied post-pooling to introduce non-linearity, enabling the model to capture complex patterns.
    \item \textbf{Intermediate Aggregation}: The regions of the local feature map are aggregated into representative values, preserving important information.
    \item \textbf{Reduction}: Spatial dimensions are reduced to lower computational costs and prevent overfitting.
    \item \textbf{Maximum}: Max pooling selects the most prominent features within each window, enhancing feature robustness.
    \item \textbf{Adaptive}: Pooling parameters are dynamically computed based on the target output dimensions, offering flexibility across architectures.
    \item \textbf{Localized}: Operations are performed on local regions, maintaining spatial context during downsampling.
\end{itemize}

\section{NIRMAL Pooling Algorithm}
\label{sec:algorithm}
The NIRMAL Pooling layer operates through a structured process, as outlined in Algorithm \ref{alg:nirmal_pooling}. It receives an input feature map from a preceding layer (typically convolutional) and produces a downsampled, activated feature map for subsequent layers.

\begin{algorithm}
\caption{NIRMAL Pooling Algorithm}
\label{alg:nirmal_pooling}
\begin{algorithmic}[1]
\Require Input feature map $I$ of shape $(BatchSize, H_{in}, W_{in}, C)$, desired output size $(H_{out}, W_{out})$
\Ensure Output feature map $O$ of shape $(BatchSize, H'_{out}, W'_{out}, C)$
\State Compute pooling window sizes: $P_h = \lceil \frac{H_{in}}{H_{out}} \rceil$, $P_w = \lceil \frac{W_{in}}{W_{out}} \rceil$
\State Compute strides: $S_h = \max(1, \lfloor \frac{H_{in}}{H_{out}} \rfloor)$, $S_w = \max(1, \lfloor \frac{W_{in}}{W_{out}} \rfloor)$
\State Initialize output tensor $O'$ of shape $(BatchSize, H'_{out}, W'_{out}, C)$, where $H'_{out} = \lfloor \frac{H_{in} - P_h}{S_h} \rfloor + 1$, $W'_{out} = \lfloor \frac{W_{in} - P_w}{S_w} \rfloor + 1$
\For{each channel $c \in \{1, \dots, C\}$}
    \For{each output position $(i,j) \in \{1, \dots, H'_{out}\} \times \{1, \dots, W'_{out}\}$}
        \State Define window: $W_{i,j} = \{ (x,y) \mid i \cdot S_h \leq x < i \cdot S_h + P_h, j \cdot S_w \leq y < j \cdot S_w + P_w \}$
        \State Compute $O'_{i,j,c} = \max_{(x,y) \in W_{i,j}} (I_{x,y,c})$
    \EndFor
\EndFor
\State Apply ReLU activation: $O_{i,j,c} = \max(0, O'_{i,j,c})$ for all $(i,j,c)$
\State \Return $O$
\end{algorithmic}
\end{algorithm}

\begin{figure}[H]
    \centering
    \begin{subfigure}[b]{0.35\textwidth}
        \centering
        \includegraphics[width=\textwidth]{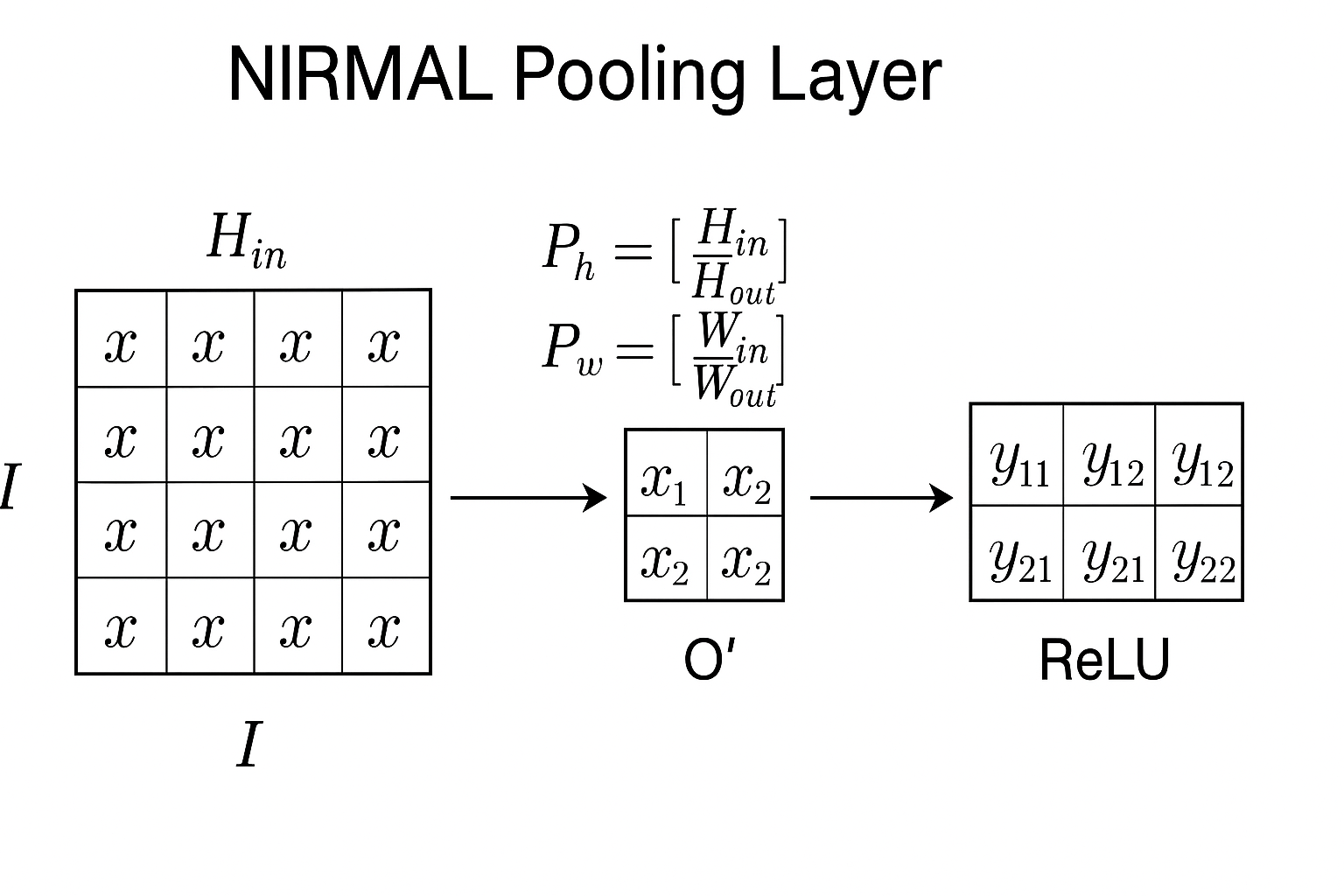}
        \caption{NIRMAL Pooling Layer}
        \label{fig:nirmal_pooling_layer}
    \end{subfigure}
    \hspace{0.5cm} 
    \begin{subfigure}[b]{0.20\textwidth}
        \centering
        \includegraphics[width=\textwidth]{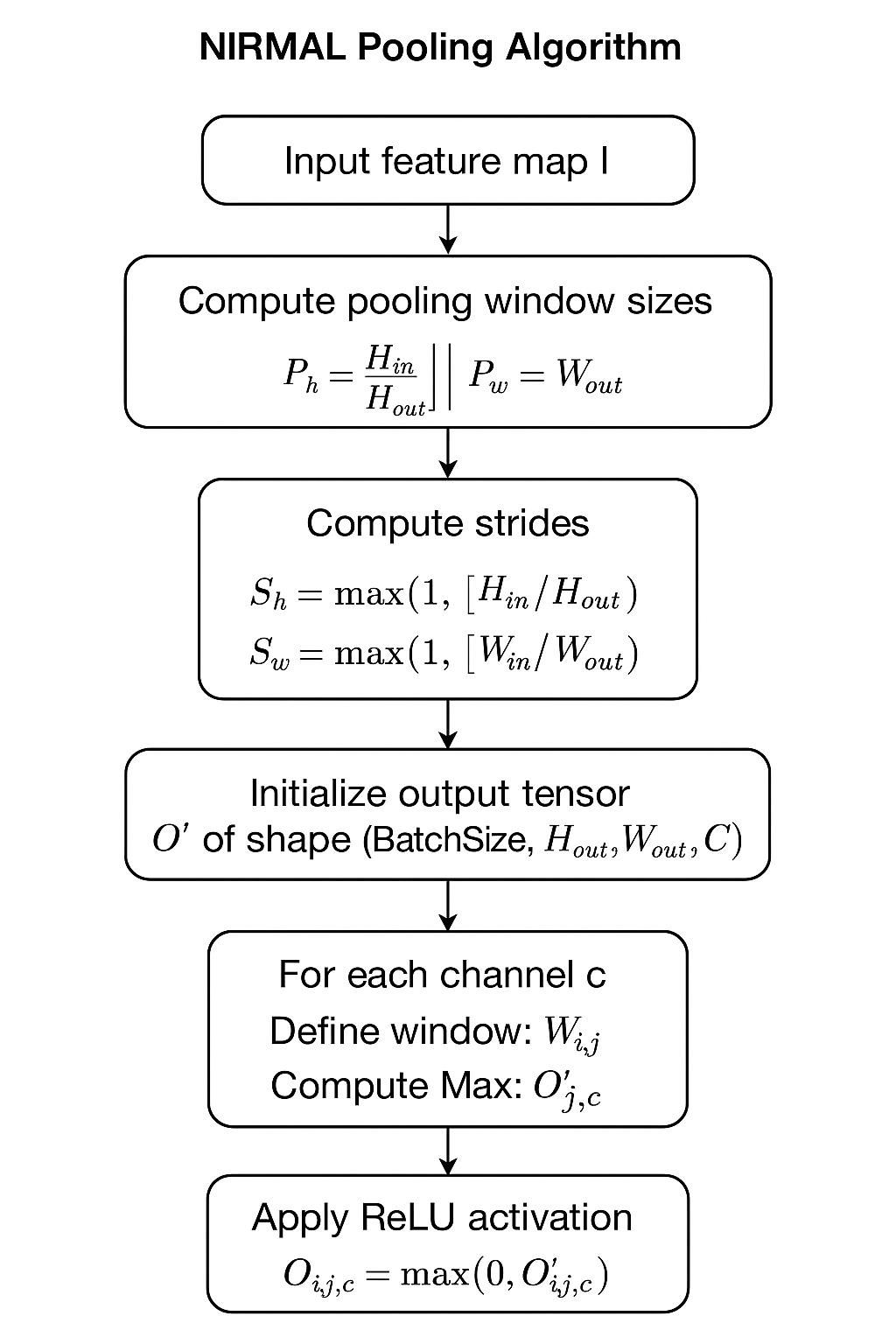}
        \caption{NIRMAL Pooling Algorithm}
        \label{fig:nirmal_pooling_algorithm}
    \end{subfigure}
    \caption{Visualizations of the NIRMAL Pooling Layer and Algorithm}
    \label{fig:nirmal_pooling_combined}
\end{figure}

The algorithm ensures that the output dimensions closely approximate the desired $H_{out} \times W_{out}$, with slight variations due to ceiling and floor operations. The integration of ReLU enhances the model’s ability to learn non-linear feature relationships, a critical factor in deep architectures.

\section{Mathematical Formulation}
\label{sec:math}
Let $I \in \mathbb{R}^{H_{in} \times W_{in} \times C}$ be the input feature map, where $H_{in}$ and $W_{in}$ are height and width, and $C$ is the number of channels. The goal is to produce an output feature map $O \in \mathbb{R}^{H'_{out} \times W'_{out} \times C}$ with target dimensions $H_{out} \times W_{out}$.

\subsection{Adaptive Parameter Calculation}
Pooling window sizes and strides are calculated as:
\begin{align}
P_h &= \left\lceil \frac{H_{in}}{H_{out}} \right\rceil, \quad P_w = \left\lceil \frac{W_{in}}{W_{out}} \right\rceil \label{eq:pool_size} \\
S_h &= \max\left(1, \left\lfloor \frac{H_{in}}{H_{out}} \right\rfloor\right), \quad S_w = \max\left(1, \left\lfloor \frac{W_{in}}{W_{out}} \right\rfloor\right) \label{eq:stride}
\end{align}
These ensure that the pooling operation adapts to the input and desired output dimensions.

\subsection{Max Pooling}
For each channel $c$ and output position $(i,j)$, the pooled value is:
\begin{equation}
O'_{i,j,c} = \max_{(x,y) \in W_{i,j}} (I_{x,y,c}), \label{eq:max_pool}
\end{equation}
where $W_{i,j} = \{ (x,y) \mid i \cdot S_h \leq x < i \cdot S_h + P_h, j \cdot S_w \leq y < j \cdot S_w + P_w \}$. The output dimensions are:
\begin{equation}
H'_{out} = \left\lfloor \frac{H_{in} - P_h}{S_h} \right\rfloor + 1, \quad W'_{out} = \left\lfloor \frac{W_{in} - P_w}{S_w} \right\rfloor + 1 \label{eq:out_size}
\end{equation}

\subsection{Non-linear Activation}
A ReLU activation is applied element-wise:
\begin{equation}
O_{i,j,c} = \max(0, O'_{i,j,c}) \label{eq:relu}
\end{equation}
This step enhances the model’s capacity to learn complex, non-linear patterns.

\section{Implementation Details}
\label{sec:implementation}
NIRMAL Pooling is implemented as a custom layer compatible with deep learning frameworks such as TensorFlow or PyTorch. The layer accepts an input tensor and a target output size, dynamically computing the pooling parameters. Key considerations include:
\begin{itemize}
    \item \textbf{Handling Edge Cases}: Ensuring minimum stride values and padding strategies to manage non-integer divisions.
    \item \textbf{Computational Efficiency}: Leveraging optimized max pooling operations from framework backends while adding minimal overhead for adaptive calculations and ReLU application.
    \item \textbf{Compatibility}: Designing the layer to integrate seamlessly with standard CNN architectures, such as LeNet, VGG, or ResNet.
\end{itemize}
The implementation is tested on GPUs to ensure scalability, with attention to numerical stability during parameter calculations.

\section{Experimental Results}
\label{sec:results}
We evaluated NIRMAL Pooling against standard Max Pooling using a consistent CNN architecture across three benchmark datasets: MNIST Digits (28$\times$28 grayscale images, 10 classes), MNIST Fashion (28$\times$28 grayscale images, 10 classes), and CIFAR-10 (32$\times$32 color images, 10 classes). The CNN architecture comprised two convolutional layers (32 and 64 filters, 3$\times$3 kernels), followed by pooling, two dense layers (128 and 10 units), and ReLU activations except for the final softmax layer. Training was conducted for 10 epochs with a batch size of 64, a 10\% validation split, and the Adam optimizer.

\begin{table}[H]
\centering
\caption{Test Performance Comparison Across Datasets}
\label{tab:results_summary}
\begin{adjustbox}{width=0.65\textwidth}
\begin{tabular}{lcccc}
\toprule
\textbf{Dataset} & \multicolumn{2}{c}{\textbf{Max Pooling}} & \multicolumn{2}{c}{\textbf{NIRMAL Pooling}} \\
\cmidrule(lr){2-3} \cmidrule(lr){4-5}
& \textbf{Loss} & \textbf{Accuracy} & \textbf{Loss} & \textbf{Accuracy} \\
\midrule
MNIST Digits  & 0.0299 & 99.12\% & 0.0238 & 99.25\% \\
MNIST Fashion & 0.2378 & 91.44\% & 0.2367 & 91.59\% \\
CIFAR-10      & 0.9032 & 68.87\% & 0.8458 & 70.49\% \\
\bottomrule
\end{tabular}
\end{adjustbox}
\end{table}

\subsection{MNIST Digits}
NIRMAL Pooling achieved a test accuracy of 99.25\% and a test loss of 0.0238, compared to 99.12\% and 0.0299 for Max Pooling. The marginal improvement suggests that both methods perform well on this relatively simple dataset, but NIRMAL’s non-linear activation provides a slight edge.

\subsection{MNIST Fashion}
In MNIST Fashion, NIRMAL Pooling yielded a test accuracy of 91.59\% and a test loss of 0.2367, compared to 91.44\% and 0.2378 for Max Pooling. The consistent improvement indicates that the adaptive nature of NIRMAL Pooling enhances performance on slightly more complex data.

\subsection{CIFAR-10}
The most significant improvement was observed on CIFAR-10, where NIRMAL Pooling achieved a test accuracy of 70.49\% and a test loss of 0.8458, compared to 68.87\% and 0.9032 for Max Pooling. This suggests that NIRMAL Pooling’s design is particularly effective for complex, color image datasets, likely due to its ability to adaptively capture salient features and enhance them with non-linear activation.

\section{Discussion}
\label{sec:discussion}
The experimental results demonstrate that NIRMAL Pooling consistently outperforms standard Max Pooling across all datasets, with the most pronounced improvement on CIFAR-10. This can be attributed to two key factors:
\begin{itemize}
    \item \textbf{Adaptive Parameterization}: By dynamically adjusting window sizes and strides, NIRMAL Pooling better accommodates varying input dimensions and task complexities, unlike the fixed parameters of Max Pooling.
    \item \textbf{Non-linear Activation}: The integration of ReLU post-pooling introduces non-linearity at an earlier stage, enabling the model to learn more complex feature hierarchies, which is particularly beneficial for datasets with richer visual content like CIFAR-10.
\end{itemize}

The marginal improvements on MNIST Digits and Fashion suggest that simpler datasets may not fully exploit the capabilities of NIRMAL Pooling, as their feature spaces are less complex. However, significant gains in CIFAR-10 highlight its potential for real-world applications involving diverse and challenging image data.

Limitations include potential computational overhead from dynamic parameter calculations and the need for careful tuning of target output dimensions to avoid information loss. Future implementations could optimize these aspects to improve scalability.

\section{Future Work}
\label{sec:future}
Several directions can further enhance the applicability of NIRMAL Pooling:
\begin{itemize}
    \item \textbf{Alternative Activation Functions}: Investigating functions like Leaky ReLU, ELU, or Swish to optimize non-linear feature enhancement.
    \item \textbf{Learnable Parameters}: Introducing trainable parameters for pooling operations to allow the model to optimize window sizes or strides during training.
    \item \textbf{Extended Comparisons}: Comparing NIRMAL Pooling with other adaptive pooling methods, such as Global Average Pooling or Spatial Pyramid Pooling \citep{he2015spatial}.
    \item \textbf{Deeper Architectures}: Testing NIRMAL Pooling in advanced architectures like ResNet or EfficientNet on large-scale datasets like ImageNet.
    \item \textbf{Computational Optimization}: Reducing the overhead of adaptive calculations through optimized implementations or hardware acceleration.
    \item \textbf{Theoretical Analysis}: Conducting a deeper analysis of NIRMAL Pooling’s impact on feature invariance and model convergence properties.
\end{itemize}

\section{Conclusion}
NIRMAL Pooling introduces a novel approach to pooling in CNNs by combining adaptive max pooling with non-linear activation. Its ability to dynamically adjust pooling parameters and enhance feature representation through ReLU activation results in consistent performance improvements over standard Max Pooling, particularly on complex datasets like CIFAR-10. The proposed method offers a flexible and robust alternative for modern CNN architectures, with potential applications in various computer vision tasks. Future work will focus on optimizing its computational efficiency and exploring its scalability in deeper networks and larger datasets.

\bibliographystyle{plainnat}
\bibliography{references}

\begin{thebibliography}{7}
\providecommand{\natexlab}[1]{#1}
\providecommand{\url}[1]{\texttt{#1}}
\expandafter\ifx\csname urlstyle\endcsname\relax
  \providecommand{\doi}[1]{doi: #1}\else
  \providecommand{\doi}{doi: \begingroup \urlstyle{rm}\Url}\fi

\bibitem[Akgül(2025)]{akgul2025mam}
İ. Akgül.
\newblock A pooling method developed for use in convolutional neural networks.
\newblock \emph{Scientific Reports}, 2025.

\bibitem[He et~al.(2015)He, Zhang, Ren, and Sun]{he2015spatial}
Kaiming He, Xiangyu Zhang, Shaoqing Ren, and Jian Sun.
\newblock Spatial pyramid pooling in deep convolutional networks for visual recognition.
\newblock \emph{IEEE Transactions on Pattern Analysis and Machine Intelligence}, 37\penalty0 (9):\penalty0 1904--1916, 2015.

\bibitem[Krizhevsky et~al.(2012)Krizhevsky, Sutskever, and Hinton]{krizhevsky2012imagenet}
Alex Krizhevsky, Ilya Sutskever, and Geoffrey~E. Hinton.
\newblock Imagenet classification with deep convolutional neural networks.
\newblock \emph{Advances in Neural Information Processing Systems}, 25, 2012.

\bibitem[Rodriguez-Martinez et~al.(2022)]{rodriguez2022pooling}
R.~Rodriguez-Martinez et~al.
\newblock A study on the suitability of different pooling operators for convolutional neural networks in the prediction of covid-19 through chest x-ray image analysis.
\newblock \emph{Applied Sciences}, 2022.

\bibitem[Shadoul et~al.(2022)]{shadoul2022effect}
I.~Shadoul et~al.
\newblock The effect of pooling parameters on the performance of convolution neural network.
\newblock \emph{Artificial Intelligence Review}, 2022.

\bibitem[Zafar et~al.(2024)]{zafar2024comparison}
A.~Zafar et~al.
\newblock A comparison of pooling methods for convolutional neural networks.
\newblock \emph{Expert Systems with Applications}, 2024.

\bibitem[Zhao and Zhang(2022)]{zhao2022improved}
L.~Zhao and Z.~Zhang.
\newblock An improved pooling method for convolutional neural networks.
\newblock 2022.

\end{thebibliography}
\end{document}